# A Hybrid Algorithm for Convex Semidefinite Optimization

**Sören Laue**                                                    SOEREN.LAUE@UNI-JENA.DE
Friedrich-Schiller-University Jena, Germany

## Abstract

We present a hybrid algorithm for optimizing a convex, smooth function over the cone of positive semidefinite matrices. Our algorithm converges to the global optimal solution and can be used to solve general large-scale semidefinite programs and hence can be readily applied to a variety of machine learning problems. We show experimental results on three machine learning problems. Our approach outperforms state-of-the-art algorithms.

## 1. Introduction

We consider the following unconstrained semidefinite optimization problem:

$$\begin{aligned} \min \quad & f(X) \\ s.t. \quad & X \succeq 0 \ , \end{aligned} \qquad (1)$$

where $f(X) : \mathbb{R}^{n \times n} \to \mathbb{R}$ is a convex and differentiable function over the cone of positive semidefinite matrices. Many machine learning problems can be cast as a semidefinite optimization problem. Prominent examples include sparse PCA (d'Aspremont et al., 2007), distance metric learning (Xing et al., 2002), nonlinear dimensionality reduction (Weinberger et al., 2006), multiple kernel learning (Lanckriet et al., 2004), multitask learning (Obozinski et al., 2010), and matrix completion (Srebro et al., 2004).

We provide an algorithm that solves general large-scale unconstrained semidefinite optimization problems efficiently. The idea to our algorithm is a hybrid approach: we combine the algorithm of Hazan (2008) with a standard quasi-Newton algorithm. The algorithm achieves convergence to the global optimum with very good running time. It can be readily used for a variety of machine learning problems and we demonstrate its efficiency on three different tasks: matrix completion, metric learning, and sparse PCA. Another advantage of the algorithm is its simplicity, as it can be implemented in less than 30 lines of Matlab code.

### 1.1. Related Work

A constrained version of Problem 1 is called a semidefinite program (SDP) if function $f$ as well as the constraints are linear. Semidefinite programs have gained a lot of attention in recent years, since many NP-hard problems can be relaxed into SDPs and many machine learning problems can be modeled as SDPs.

The most widely known implementations of SDP solvers are interior point methods. They provide high-accuracy solutions in polynomial time. However, since the running time is a low-order polynomial in the dimension $n$ they do not scale well to medium and large problems that often occur in machine learning. On the other hand, the high accuracy of their solutions is typically not needed as the input data is often noisy. Among other methods, proximal methods have been employed to solve SDPs in order to circumvent the large running time of interior point methods. They achieve better running times at the expense of less accurate solutions. Examples include (Nesterov, 2007; Nemirovski, 2004) and (Arora et al., 2005) where the multiplicative weights update rule is employed.

The algorithm of (Arora et al., 2005) has been randomized by (Garber & Hazan, 2011) based on the same idea as in (Grigoriadis & Khachiyan, 1995) to achieve sublinear running time. Another randomized algorithm has appeared in (Kleiner et al., 2010). Furthermore, alternating direction methods have been proposed to solve SDPs (Wen et al., 2010).

Another line of algorithms for solving SDPs are Frank-Wolfe type algorithms such as (Hazan, 2008). This approach is also known as sparse greedy approximation and these algorithms have the advantage that they produce sparse solutions (Clarkson, 2008) which, for SDPs, corresponds to low-rank solutions. Low-rank solutions are very appealing since they can drastically reduce the computational effort. Instead of storing a low-rank positive semidefinite matrix $X \in \mathbb{R}^{n \times n}$ one just stores a matrix $V \in \mathbb{R}^{n \times k}$ where $X = VV^T$, with $k$ being the rank of $X$. Matrix-vector multiplications, for instance, can be done in $O(nk)$ instead of $O(n^2)$.

There have also been nonlinear approaches to linear SDPs, however, without general convergence guarantees (Burer & Monteiro, 2003). In some special cases (matrix completion problems where it is assumed that the data is indeed generated by a low rank matrix and the restricted isometry property holds) convergence guarantees have been shown. In general, the problem of solving an SDP with a low-rank constraint is NP-hard (Goemans & Williamson, 1995).

**Organization of the Paper** Section 2 describes our algorithm while in Section 3 we provide experimental results on three different machine learning problems. We compare our algorithm against a standard interior point method and against algorithms that are specifically designed for each of the individual problems. Section 4 provides theoretical guarantees for the running time and for the convergence to the global optimal solution.

## 2. The Hybrid Algorithm

Our algorithm is summarized in pseudo-code in Algorithm 1.

---

**Algorithm 1** Hybrid Algorithm

---

**Input:** Smooth, convex function $f : \mathbb{R}^{n \times n} \to \mathbb{R}$
**Output:** Approximate solution of Problem 1
Initialize $X_0 = 0$, $V_0 = 0$
**repeat**
 Increase rank by 1 using Hazan update:
  Compute $v_i = \text{ApproxEV}(-\nabla f(V_i V_i^T), \tilde{\varepsilon})$.
  Solve

$$\min_{\alpha, \beta} \quad f(\alpha \cdot V_i V_i^T + \beta \cdot v_i v_i^T)$$
$$\text{s.t.} \quad \alpha, \beta \geq 0.$$

  Set $V_{i+1} = [\sqrt{\alpha} \cdot V_i, \sqrt{\beta} \cdot v_i]$.
 Run nonlinear update:
  Improve $V_{i+1}$ by finding a local minimum
  of $f(VV^T)$ wrt. $V$ starting with $V_{i+1}$.
**until** Approximation guarantee has been reached

---

The notation $[V_i, v_i]$ used in Algorithm 1 stands for the horizontal concatenation of matrix $V_i$ and column vector $v_i$. The function ApproxEV returns an approximate eigenvector to the largest eigenvalue:

given a square matrix $M$ it returns a vector $v_i = \text{ApproxEV}(M, \tilde{\varepsilon})$ of unit length that satisfies $v_i^T M v_i \geq \lambda_{\max}(M) - \tilde{\varepsilon}$, where $\lambda_{\max}(M)$ denotes the largest eigenvector of matrix $M$.

Our algorithm runs in iterations. Each iteration consists of two steps: a rank-1 update and a subsequent nonlinear improvement of the current solution. The rank-1 update step follows a Frank-Wolfe type approach. A linear approximation to function $f$ at the current iterate $X_i$ is minimized over the cone of semidefinite matrices. The minimum is attained at $v_i v_i^T$ where $v_i = \lambda_{\max}(-\nabla f(X_i))$ is the vector to the largest eigenvalue of $-\nabla f(X_i)$. Then the next iterate $X_{i+1}$ is a linear combination of the current iterate $X_i = V_i V_i^T$ and $v_i v_i^T$ such that it minimizes $f$.

In the second step, the nonlinear update step, the current solution $X_{i+1} = V_{i+1} V_{i+1}^T$ is further improved by minimizing function $f(VV^T)$ with respect to $V$. Note that this function is no longer convex with respect to $V$. Hence, we can only expect to find a local minimum. Our analysis however shows, that this is sufficient to still converge to the global optimal solution of Problem 1. In fact, it is even not necessary to find a local minimum, any improvement will work.

In Section 4 we will prove that after at most $O(\frac{1}{\varepsilon})$ many iterations Algorithm 1 will return a solution that is $\varepsilon$-close to the global optimal solution.

## 3. Applications and Experiments

We have implemented our hybrid algorithm in Matlab exactly as described in Algorithm 1. For the nonlinear update we use `minFunc` (Schmidt) which implements the limited memory BFGS algorithm. The two-variable optimization problem in the rank-1 update is also solved using `minFunc`. We use the default settings of `minFunc`. The approximate eigenvector computation is done using the Matlab function `eigs`. We ran all experiments in single-thread mode on a 2.50GHz CPU.

### 3.1. Matrix Completion

In this section we consider the matrix completion problem used for collaborative filtering. Given a matrix $Y$ where only a few entries have been observed the goal is to complete this matrix by finding a low-complexity matrix $X$ which approximates the given entries of $Y$ as good as possible. Low complexity can be achieved for instance by a low rank or by small trace norm. Also different error norms can be considered depending on the specific application. For instance, the l1-norm in combination with the trace-norm regularization leads to the robust PCA approach for matrix completion with low rank (Candès et al., 2011). Here, we use the l2-norm and the rank constraint as a measure of complexity. Hence, the matrix completion problem becomes the following optimization problem:

$$\min \quad \sum_{(i,j) \in \Omega} (X_{ij} - Y_{ij})^2$$
$$s.t. \quad \mathrm{rank}(X) \leq k \ , \tag{2}$$

where $\Omega$ is the set of all given entries of $Y$. Note that Problem (2) is NP-hard (Gillis & Glineur, 2011). However, we can still attempt to find a good solution to it by using our hybrid algorithm. Problem (2) can be transformed into the following equivalent semidefinite optimization problem:

$$\min \quad \sum_{(i,j) \in \hat{\Omega}} (\hat{X}_{ij} - \hat{Y}_{ij})^2$$
$$s.t. \quad \mathrm{rank}(\hat{X}) \leq k \tag{3}$$
$$\hat{X} \succeq 0 \ ,$$

where

$$\hat{X} = \begin{pmatrix} V & X \\ X^T & W \end{pmatrix} \text{ and } \hat{Y} = \begin{pmatrix} 0 & Y \\ Y^T & 0 \end{pmatrix},$$

and $\hat{X}$ is a positive semidefinite matrix. $V$ and $W$ are suitable symmetric matrices. Hence, the matrix completion Problem (2) for an input matrix $Y \in \mathbb{R}^{m \times n}$ can be cast into a semidefinite optimization problem over matrices $X \in \mathbb{R}^{(m+n) \times (m+n)}$.

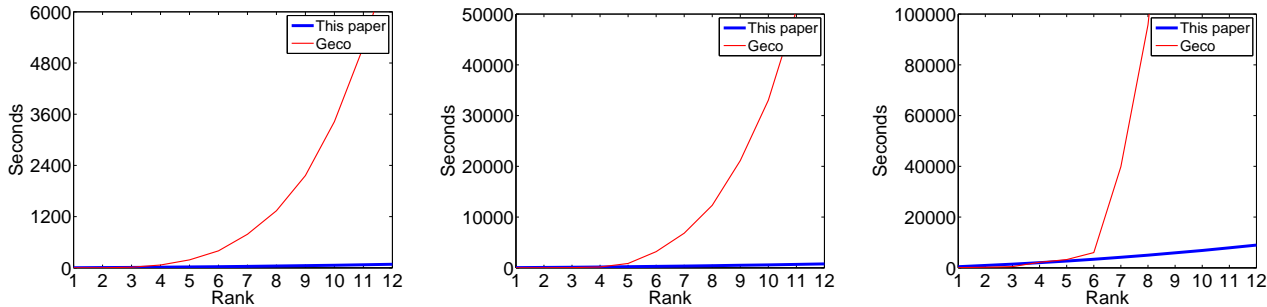

*Figure 1.* Matrix completion problem: running time with respect to the rank for the three MovieLens datasets: 100K, 1M, and 10M.

*Table 1.* Matrix completion problem: test error (RMSE) on the three MovieLens datasets.

|  | 100K | | | 1M | | | 10M | | |
|---|---|---|---|---|---|---|---|---|---|
|  | RMSE | Sec. | Rank | RMSE | Sec. | Rank | RMSE | Sec. | Rank |
| GECO | 0.947 | 397 | 6 | 0.874 | 50712 | 11 | 0.821 | 784941 | 12 |
| GECO | 0.950 | 127 | 5 | 0.880 | 6368 | 7 | 0.830 | 65468 | 8 |
| THIS PAPER | 0.933 | 5 | 2 | 0.874 | 223 | 5 | 0.815 | 2663 | 5 |

We compare our algorithm to a state-of-the-art solver GECO (Shalev-Shwartz et al., 2011) which was specifi-cally designed for solving large-scale matrix minimization problems with a low-rank constraint. We follow the experimental setting of (Shalev-Shwartz et al., 2011). We use three standard matrix completion datasets: MovieLens100k, MovieLens1M, and MovieLens10M. The dimensions of the three datasets are $943 \times 1682$, $6040 \times 3706$, and $69878 \times 10677$ respectively and they contain $10^5$, $10^6$, and $10^7$ movie ratings from 1 to 5. The task is to predict a movie rating for user $i$ and movie $j$. We used the datasets without any normalization[1] and split them randomly such that for each user 80% of the ratings went into training data and 20% into test data.

Our algorithm minimizes the training error much faster than GECO and at the same time also needs a much smaller rank. As a result we also achieve an optimal *test* error much faster and with a smaller rank than GECO. Table 1 reports the test root-mean-square error (RMSE) as well as the rank where it was achieved and the running times. Both rows for GECO in Table 1 reflect the same runs. The first row shows the statistics where the test error reaches the minimum. However, since GECO slows down a lot with the rank we also added intermediate results when the test error is approaching the minimum. As it can be observed our algorithm achieves the same or better test error by requiring only a fraction of the time needed by GECO.

---

[1] We noticed that normalization had no impact on the results.

Both algorithms need quasi-linear time in the number of ratings and hence can be used to solve large scale matrix factorization problems. Note however, that for our algorithm the runtime per iteration scales linearly with the rank $k$ whereas GECO needs $O(k^6)$. This behavior slows down GECO considerably, which can be seen in Figure 1.

## 3.2. Metric Learning

The second problem we approach is the metric learning problem. We are given a labeled dataset $X = (x_i, y_i)_i$. Let $S$ be a set containing all pairs of indices $(i,j)$ whose data points $x_i$ and $x_j$ are similar to each other, i.e. its labels $y_i$ and $y_j$ are equal; let the set $\bar{S}$ contain all indices of data points that are dissimilar to each other. For a given semidefinite matrix $A$, the Mahalanobis distance between $x_i$ and $x_j$ is defined as $d_A(i,j) = \sqrt{(x_i - x_j)^T A (x_i - x_j)}$. The metric learning problem is that of finding a positive semidefinite matrix $A$, such that under the induced Mahalanobis distance, points that are similar are close to each other and points that are dissimilar are far apart. This problem can be cast as a semidefinite optimization problem(Xing et al., 2002):

$$\begin{aligned} \min \quad & \textstyle\sum_{(i,j) \in S} d_A(i,j)^2 \\ s.t. \quad & \textstyle\sum_{(i,j) \in \bar{S}} d_A(i,j) \geq 1 \\ & A \succeq 0. \end{aligned} \quad (4)$$

Note that the 1 in the inequality constraint in Problem (4) can be changed to any arbitrary positive constant. This constraint is just to ensure that not all points are mapped onto the same point. Problem (4) does not fit into our framework. However, we can transform it into the following equivalent unconstrained semidefinite problem:

$$\begin{aligned} \min \quad & \textstyle\sum_{(i,j) \in S} d_A(i,j)^2 - \lambda \sum_{(i,j) \in \bar{S}} d_A(i,j) \\ \text{s. t.} \quad & A \succeq 0. \end{aligned} \quad (5)$$

Problem (5) is just the Lagrangian of Problem (4). Since the 1 in the inequality constrained was chosen arbitrary we can also choose any positive constant for $\lambda$. In our experiments we set it to 1.

We follow the experimental setting of (Kleiner et al., 2010). We compared our approach against an interior point method implemented in SeDuMi (Sturm, 1999) (via CVX (Grant & Boyd, 2011)) and the algorithm of (Xing et al., 2002) which is a projected gradient approach and was specifically designed to solve the above SDP. We could not directly compare our algorithm to that of (Kleiner et al., 2010) as the code was not available. However, the authors show that it performs similarly to (Xing et al., 2002).

As a measure of quality for a given solution $A$ we define:

$$Q(A) = \frac{1}{\xi} \cdot \sum_i \sum_{j:(i,j) \in S} \sum_{l:(i,l) \in \bar{S}} \mathbb{1}[d_A(i,j) < d_A(i,l)]),$$

where $\mathbb{1}[.]$ is the indicator function and $\xi = \sum_i \sum_{j:(i,j) \in S} \sum_{l:(i,l) \in \bar{S}} 1$ is a normalization factor. In essence $Q$ captures how many points with the same label are mapped closer to each other than points with different labels.

We initially apply metric learning to the UCI ionosphere dataset which contains 351 labeled data points in dimension 34. The results of this experiment are shown in Figure 2. As it can be seen, our hybrid algorithm achieves the optimal value almost instantly. The projected gradient descent algorithm (PG) needs about 20 times as long to achieve a solution of comparable quality. Since the interior point method (IP) scales very badly with the number of data points, we only ran it on a sub-sample of size 4*34=136. On this dataset, our method achieves the same function value as IP: 5.47e-05, while requiring only 0.28 seconds as opposed to 1513 seconds that IP needs. PG achieves a function value of 5.50e-05 in 88 seconds.

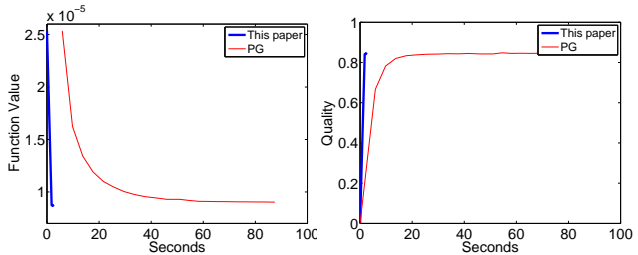

*Figure 2.* Metric learning problem: UCI ionosphere data.

Following (Kleiner et al., 2010), we ran a second set of experiments on synthetic data in order to measure the dependence on the dimension $d$. We sampled points from $\mathbb{R}^d$ as follows: We define two sets of cluster centers $C_1 = \{(-1, 1), (-1, -1)\}$ and $C_2 = \{(1, -1), (1, 1)\}$ and apply a random rotation to both sets. We then sample each data point from a uniform Gaussian distribution $\mathcal{N}(0, I_d)$. The first two coordinates of each data point are replaced by one of the cluster centers and the label of this data point is set accordingly to either 1 or 2. Finally, a small perturbation drawn from $\mathcal{N}(0, 0.25I_2)$ is added to the first two coordinates. The results are depicted in Table 2, which shows the running times for the various algorithms until a quality measure of $Q > 0.99$ has been reached.

Our algorithm achieves the same optimal function values as the interior point method, while requiring less time than the PG method. For larger dimensions we plot the results in Figure 3. As it can be observed, our algorithm is considerably faster than PG on these larger dimensions. We omit the IP method here, as this scales badly with increased dimension.

*Table 2.* Metric learning problem: synthetic data. Time to $Q > 0.99$ for different dimensions and function values at those times.

| $d$ | Alg. | $f$ value | Sec. to Q>0.99 |
|---|---|---|---|
| 50 | PG | 3.94e-05 | 1.44 |
| 50 | IP | 3.57e-06 | 3713 |
| 50 | THIS PAPER | 3.37e-06 | 1.39 |
| 100 | PG | 5.62e-05 | 8.48 |
| 100 | IP | 1.21e-06 | 9048 |
| 100 | THIS PAPER | 1.19e-06 | 2.34 |

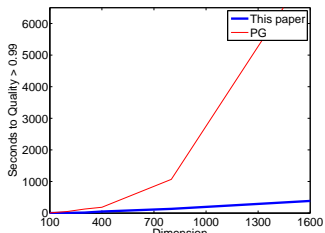

*Figure 3.* Metric learning problem: Synthetic data.

### 3.3. Sparse PCA

As a third problem we consider the sparse principal component analysis problem (sparse PCA). For a given covariance matrix $A \in \mathbb{R}^{n \times n}$, sparse PCA tries to find a sparse vector $x$ that maximizes $x^T A x$, i.e. a sparse principal component of $A$. This problem can be relaxed into the following SDP (d'Aspremont et al., 2007):

$$\begin{aligned} \min \quad & \rho \sum_{(i,j)} |X_{ij}| - A \bullet X \\ \text{s.t.} \quad & \text{Tr}(X) = 1 \\ & X \succeq 0, \end{aligned} \quad (6)$$

where $A \bullet X$ denotes $\text{Tr}(A^T X)$. In a subsequent rounding step the largest eigenvector of the solution to Problem (6) is returned as the solution vector $x$. The parameter $\rho$ controls the tradeoff between the sparsity of $x$ and the explained variance $x^T A x$.

Problem (6) is not in form (1). However, one can easily transform it into an unconstrained semidefinite problem by defining the functions $g(X) = \frac{X}{\text{Tr}(X)}$ and $f(X) = \rho \sum_{(i,j)} |X_{ij}| - A \bullet X$. Hence, Problem (6) is equivalent to

$$\begin{aligned} \min \quad & f(g(X)) \\ & X \succeq 0. \end{aligned} \quad (7)$$

Note that $f(g(X))$ is again a convex function over the set of semidefinite matrices without the zero matrix. However, $f(g(X))$ is not smooth. Smoothness of $f(g(X))$ can be achieved either by implicitly smoothing it, e.g. using Nesterov's smoothing technique (Nesterov, 2005) or by explicitly smoothing it and replacing the absolute function $|.|$ with the scaled Huber-loss $\mathcal{H}_M$. The Huber-loss is defined as:

$$\mathcal{H}_M(x) = \begin{cases} x^2 & \text{if } |x| \le M \\ 2M|x| - M^2 & \text{if } |x| > M \end{cases}$$

By appropriate scaling one can achieve an arbitrary small difference between $|x|$ and $\mathcal{H}_M(x)$. We obtain a smooth, convex function $f_{\mathcal{H}_M}$ by replacing the absolute function with the Huber-loss in function $f$. In our experiments we set $M = 10^{-6}$ such that functions $f_{\mathcal{H}_M}$ and $f$ differ only marginally from each other.

We again follow the experimental setting of (Kleiner et al., 2010) and we compare to an interior point method and to a state-of-the-art algorithm, the DSPCA algorithm (d'Aspremont et al., 2007) which is specifically designed to solve Problem (6). We used the colon cancer data set which contains 2000 microarray readings from 62 subjects. We randomly sampled readings in order to vary the dimension $d$. As standard with this task, we normalized the data to mean 0 and standard deviation 1. We set $\rho = 0.2$ in Problem (6) to obtain sparse solutions for all $d$.

*Table 3.* Sparse PCA problem: colon cancer data. Function value at convergence, time to convergence, sparsity and variance of the largest eigenvector $x$ of the solution $X$.

| $d$ | Alg. | $f$ value | Sec. | Spars. | Var. |
|---|---|---|---|---|---|
| 50 | DSPCA | -1.8326 | 0.22 | 0.50 | 2.53 |
| 50 | IP | -1.8675 | 140.5 | 0.88 | 2.53 |
| 50 | THIS PAPER | -1.8675 | 0.50 | 0.88 | 2.53 |
| 100 | DSPCA | -4.9235 | 1.27 | 0.81 | 6.30 |
| 100 | IP | -4.9550 | 11215 | 0.88 | 6.29 |
| 100 | THIS PAPER | -4.9550 | 0.72 | 0.88 | 6.29 |
| 200 | DSPCA | -6.27 | 9.65 | 0.82 | 8.06 |
| 200 | THIS PAPER | -6.35 | 3.31 | 0.89 | 8.05 |
| 400 | DSPCA | -16.34 | 64.29 | 0.84 | 19.81 |
| 400 | THIS PAPER | -16.48 | 19.91 | 0.87 | 19.78 |
| 800 | DSPCA | -31.74 | 595 | 0.86 | 38.40 |
| 800 | THIS PAPER | -32.00 | 99 | 0.87 | 38.37 |
| 1600 | DSPCA | -74.14 | 10034 | 0.86 | 81.29 |
| 1600 | THIS PAPER | -74.50 | 464 | 0.88 | 81.18 |

Table 3 reports the running time, the function value at convergence, the sparsity of the solution and the captured variance for these data sets. As mentioned above, we ran our algorithm on the function $f_{\mathcal{H}_M}$, however we report the function value $f(X)$ for the original formulation (6). The solutions of our algorithm are basically identical to those of the interior point method. However, it needs only a fraction of the time spent by the interior point method. The DSPCA algorithm provides accurate solutions within short time even with increasing dimension, however our algorithm is still considerably faster, especially for large dimensions.

## 4. Analysis

### 4.1. The Duality Gap

In this section we provide a duality gap for Problem (1) and analyze the running time of Algorithm 1. Let Problem (1) have finite optimal solution denoted by $f^*$ obtained at $X^*$. Let $t$ be an upper bound on the trace norm $\mathrm{Tr}(X^*)$. Such a trace bound always exists if $f^* > -\infty$. Then the optimization Problem (1) is equivalent to:

$$
\begin{aligned}
\min \quad & f(X) \\
\mathrm{s.\,t.} \quad & \mathrm{Tr}(X) \leq t \\
& X \succeq 0
\end{aligned} \tag{8}
$$

In order to simplify some technicalities in the proof we change Problem (8) into:

$$
\begin{aligned}
\min \quad & \hat{f}(\hat{X}) \\
\mathrm{s.\,t.} \quad & \mathrm{Tr}(\hat{X}) = t \\
& \hat{X} \succeq 0.
\end{aligned} \tag{9}
$$

Problems (8) and Problem (9) are equivalent if we define

$$
\hat{f}(\hat{X}) := f(X),
$$

where

$$
\hat{X} = \left( \begin{array}{cc} X & 0 \\ 0 & t' \end{array} \right)
$$

and $t' = t - \mathrm{Tr}(X) \geq 0$. Note that $\hat{X}$ is positive semidefinite whenever $X$ is positive semidefinite. This transformation is only done here for simplifying the analysis of Algorithm 1. It does not alter Algorithm 1.

We denote by $S_t := \{ X \in \mathbb{R}^{n \times n} \mid X \succeq 0, \ \mathrm{Tr}(X) = t \}$ the set of all positive semidefinite matrices with trace constraint $t$.

Hence, we have that Problem (1) is equivalent to

$$
\min f(X), \mathrm{s.\,t.}\, X \in S_t. \tag{10}
$$

By convexity of $f$, we have the following linearization, for any $X, Y \in S_t$:

$$
f(Y) \geq \nabla f(X) \bullet (Y - X) + f(X).
$$

This allows us to define the Wolfe-dual of (10) for any fixed matrix $X \in S_t$ as follows,

$$
\begin{aligned}
\omega(X) \quad := \quad & \min_{Y \in S_t} \nabla f(X) \bullet (Y - X) + f(X) \\
= \quad & f(X) - \max_{Y \in S_t} -\nabla f(X) \bullet (Y - X)
\end{aligned}
$$

and the duality gap as

$$
\begin{aligned}
g(X) \quad := \quad & f(X) - \omega(X) \\
= \quad & \max_{Y \in S_t} -\nabla f(X) \bullet (Y - X).
\end{aligned}
$$

By the definition of the objective function $f$, the gradient $\nabla f(X)$ is always a symmetric matrix and therefore has real eigenvalues, which will be important in the following.

**Lemma 1.** *The duality gap can be written as*

$$
g(X) = t \cdot \lambda_{\max} \left( -\nabla f(X) \right) + \nabla f(X) \bullet X.
$$

*Proof.* We will prove the claim by showing that for any symmetric matrix $G \in \mathbb{R}^{n \times n}$, one can equivalently reformulate the linear optimization problem $\max_{Y \in S_t} G \bullet Y$ as follows:

$$
\begin{aligned}
\max_{Y \in S_t} G \bullet Y \quad = \quad & \max\ G \bullet \sum_{i=1}^{n} \alpha_i u_i u_i^T \\
= \quad & \max \sum_{i=1}^{n} \alpha_i (G \bullet u_i u_i^T),
\end{aligned}
$$

where the latter maximization is taken over unit vectors $u_i \in \mathbb{R}^n$, $\|u_i\| = 1$, for $1 \leq i \leq n$, and real coefficients $\alpha_i \geq 0$, with $\sum_{i=1}^{n} \alpha_i = t$.

For $Y \in S_t$ let $Y = U^T U$ be its Cholesky factorization. Let $\alpha_i$ be the squared norms of the rows of $U$, and let $u_i$ be the row vectors of $U$, scaled to unit length. From the observation $\mathrm{Tr}(Y) = \mathrm{Tr}(U^T U) = \mathrm{Tr}(UU^T) = \sum_i \alpha_i = t$ it follows that any $Y \in S_t$ can be written as a convex combination of rank-1 matrices $Y = \sum_{i=1}^{n} \alpha_i u_i u_i^T$ with unit vectors $u_i \in \mathbb{R}^n$.

It follows

$$
\begin{aligned}
\max_{Y \in S_t} G \bullet Y \quad = \quad & \max \sum_{i=1}^{n} \alpha_i (G \bullet u_i u_i^T) \\
= \quad & \max \sum_{i=1}^{n} \alpha_i u_i^T G u_i \\
= \quad & t \cdot \max_{v \in \mathbb{R}^n, \|v\|=1} G \bullet vv^T \\
= \quad & t \cdot \max_{v \in \mathbb{R}^n, \|v\|=1} v^T G v \\
= \quad & t \cdot \lambda_{\max} (G) \,,
\end{aligned}
$$

where the last equality is the variational characterization of the largest eigenvalue.

Finally, both claims follow by plugging in $-\nabla f(X)$ for $G$. $\qquad\square$

By construction the duality gap $g(X)$ is always an upper bound on the primal error $h(X) = f(X) - f(X^*)$. This can also be used as a stopping criterion in Algorithm 1. If $g(X) \le \varepsilon$ then $f(X)$ is an $\varepsilon$-approximation to the optimal solution.

## 4.2. Runtime and Convergence Analysis

In this section we will show that after at most $O(\frac{1}{\varepsilon})$ iterations Algorithm 1 returns a solution that is an $\varepsilon$-approximation to the global optimal solution. The proof is along the lines of (Clarkson, 2008) and (Hazan, 2008). However, we improve by lowering the needed accuracy for the eigenvector computation from $O(\varepsilon^2)$ to $O(\varepsilon)$. This in turn lowers the computational effort for a eigenvector computation from $O(\frac{1}{\varepsilon})$ to $O(\frac{1}{\sqrt{\varepsilon}})$ per iteration when using the Lanczos method.

Let the curvature constant $C_f$ be defined as follows:

$$C_f :=$$
$$\sup_{\substack{X, Z \in S_t, \alpha \in [0,1] \\ Y = X + \alpha(Z - X)}} \frac{1}{\alpha^2} \left( f(Y) - f(X) - (Y - X) \bullet \nabla f(X) \right).$$

The curvature constant is a measure of how much the function $f(X)$ deviates from a linear approximation in $X$, and hence can be seen as an upper bound on the relative Bregman divergence induced by $f$. Now we can prove the following theorem.

**Theorem 2.** *For each* $i \ge 1$*, the iterate* $X_i$ *of Algorithm 1 satisfies* $f(X_i) - f(X^*) \le \varepsilon$*, where* $f(X^*)$ *is the optimal value for the minimization Problem (1), and* $\varepsilon = \frac{8C_f}{i+2}$*.*

*Proof.* We have $X_i = V_i V_i^T$. Let the sequence $\alpha_i = \frac{2}{i+2}$. For each iteration of Hazan's rank-1 update, we have that

$$f(X_{i+1})$$
$$= \min_{\alpha, \beta \ge 0} f(\alpha \cdot V_i V_i^T + \beta \cdot v_i v_i^T)$$
$$\le f((1 - \alpha_i) \cdot V_i V_i^T + \alpha_i t \cdot v_i v_i^T)$$
$$= f(X_i + \alpha_i (t \cdot v_i v_i^T - X_i))$$
$$\le f(X_i) + \alpha_i (t \cdot v_i v_i^T - X_i) \bullet \nabla f(X_i) + \alpha_i^2 C_f \quad (11)$$

where the first inequality follows from choosing $\alpha = 1 - \alpha_i$ and $\beta = \alpha_i \cdot t$ and the last inequality follows from the definition of the curvature constant $C_f$. Furthermore,

$$(t \cdot v_i v_i^T - X) \bullet \nabla f(X_i)$$
$$= (X_i - t \cdot v_i v_i^T) \bullet (-\nabla f(X_i))$$
$$= X_i \bullet (-\nabla f(X_i)) - t \cdot v_i^T (-\nabla f(X_i)) v_i$$
$$\le -X_i \bullet \nabla f(X_i) - t \cdot (\lambda_{\max}(-\nabla f(X_i)) - \tilde{\varepsilon})$$
$$\le -g(X_i) + t \cdot \tilde{\varepsilon}$$
$$\le -g(X_i) + \alpha_i \cdot C_f .$$

The last inequality follows from setting $\tilde{\varepsilon}$ to a value at most $\frac{\alpha_i \cdot C_f}{t}$ within Algorithm 1. Hence, Inequality (11) evaluates to

$$f(X_{i+1}) \le f(X_i) - \alpha_i g + \alpha_i^2 C_f + \alpha_i^2 C_f$$
$$= f(X_i) - \alpha_i g(X_i) + 2\alpha_i^2 C_f . \quad (12)$$

Subtracting $f(X^*)$ on both sides of Inequality (12), and denoting the current primal error by $h(X_i) = f(X_i) - f(X^*)$, we get

$$h(X_{i+1}) \le h(X_i) - \alpha_i g(X_i) + 2\alpha_i^2 C_f , \quad (13)$$

which by using the fact that the duality gap $g(X_i)$ is always an upper bound on the primal error $h(X_i)$ gives

$$h(X_{i+1}) \le h(X_i) - \alpha_i h(X_i) + 2\alpha_i^2 C_f . \quad (14)$$

The claim of this theorem is that the primal error $h(X_i) = f(X_i) - f(X^*)$ is small after a sufficiently large number of iterations. Indeed, we will show by induction that $h(X_i) \le \frac{8C_f}{i+2}$. In the first iteration $(i = 0)$, we know from (14) that the claim holds, because of the choice of $\alpha_0 = 1$.

Assume now that $h(X_i) \le \frac{8C_f}{i+2}$ holds. Using $\alpha_i = \frac{2}{i+2}$ in Inequality (14) we can now bound $h(X_{i+1})$ as follows:

$$h(X_{i+1}) \le h(X_i)(1 - \alpha_i) + 2\alpha_i^2 C_f$$
$$\le \frac{8C_f}{i+2}\left(1 - \frac{2}{i+2}\right) + \frac{8C_f}{(i+2)^2}$$
$$\le \frac{8C_f}{i+2} - \frac{8C_f}{(i+2)^2}$$
$$\le \frac{8C_f}{i+1+2}.$$

So far we only considered the progress made by Algorithm 1 trough the rank-1 update. Running the nonlinear improvement on $f(V_i V_i^T)$ in each iteration only improves the primal error $h(X_i)$ in each iteration. Hence, the claim of the theorem follows. $\qquad\square$

We can set $\tilde{\varepsilon} = \frac{\varepsilon}{4t}$ throughout Algorithm 1. This will ensure $\tilde{\varepsilon} \le \frac{\alpha_i \cdot C_f}{t}$ as needed by the analysis of the algorithm.

## 5. Discussion

We have provided an algorithm that optimizes convex, smooth functions over the cone of positive semidefinite matrices. It can be readily used for a variety of machine learning problems, as many of these fall into this framework or can be equivalently transformed into such a problem. In this paper we have performed experiments on three of such problems and we have shown that our algorithm significantly outperformes state-of-the-art solvers without the need of tuning it to any of the specific tasks. Additionally, the algorithm proposed has the advantage of being very simple, and it comes with the guarantee to always converge to the global optimal solution. In the future, we plan to implement our algorithm in C++ for further speedups.

#### Acknowledgements

The author would like to thank Georgiana Dinu and Joachim Giesen for useful discussions on the topic. This work was supported by the Deutsche Forschungsgemeinschaft (DFG) under grant GI-711/3-2.

## References

Arora, Sanjeev, Hazan, Elad, and Kale, Satyen. Fast algorithms for approximate semidefinite programming using the multiplicative weights update method. In *FOCS*, 2005.

Burer, Samuel and Monteiro, Renato D.C. A nonlinear programming algorithm for solving semidefinite programs via low-rank factorization. *Math. Program.*, 95(2):329–357, 2003.

Candès, Emmanuel J., Li, Xiaodong, Ma, Yi, and Wright, John. Robust principal component analysis? *J. ACM*, 58(3):11, 2011.

Clarkson, Kenneth L. Coresets, sparse greedy approximation, and the Frank-Wolfe algorithm. In *SODA*, 2008.

d'Aspremont, Alexandre, El Ghaoui, Laurent, Jordan, Michael I., and Lanckriet, Gert R. G. A direct formulation of sparse PCA using semidefinite programming. *SIAM Review*, 49(3), 2007.

Garber, Dan and Hazan, Elad. Approximating semidefinite programs in sublinear time. In *NIPS*, 2011.

Gillis, Nicolas and Glineur, François. Low-rank matrix approximation with weights or missing data is NP-hard. *SIAM J. Matrix Analysis Applications*, 32(4):1149–1165, 2011.

Goemans, Michel X. and Williamson, David P. Improved approximation algorithms for maximum cut and satisfiability problems using semidefinite programming. *J. ACM*, 42(6):1115–1145, 1995.

Grant, Michael and Boyd, Stephen. CVX: Matlab software for disciplined convex programming, version 1.21. http://cvxr.com/cvx, April 2011.

Grigoriadis, Michael D. and Khachiyan, Leonid G. A sublinear-time randomized approximation algorithm for matrix games. *Operations Research Letters*, 18(2):53–58, 1995.

Hazan, Elad. Sparse approximate solutions to semidefinite programs. In *LATIN*, 2008.

Kleiner, Ariel, Rahimi, Ali, and Jordan, Michael I. Random conic pursuit for semidefinite programming. In *NIPS*, 2010.

Lanckriet, Gert R. G., Cristianini, Nello, Bartlett, Peter, Ghaoui, Laurent El, and Jordan, Michael I. Learning the kernel matrix with semidefinite programming. *Journal of Machine Learning Research*, 5:27–72, December 2004.

Nemirovski, Arkadi. Prox-method with rate of convergence O(1/t) for variational inequalities with lipschitz continuous monotone operators and smooth convex-concave saddle point problems. *SIAM Journal on Optimization*, 15:229–251, 2004.

Nesterov, Yurii. Smooth minimization of non-smooth functions. *Math. Program.*, 103:127–152, May 2005.

Nesterov, Yurii. Smoothing technique and its applications in semidefinite optimization. *Math. Program.*, 110(2): 245–259, 2007.

Obozinski, Guillaume, Taskar, Ben, and Jordan, Michael I. Joint covariate selection and joint subspace selection for multiple classification problems. *Statistics and Computing*, 20:231–252, April 2010.

Schmidt, Mark. minFunc. http://www.di.ens.fr/~mschmidt/Software/minFunc.html.

Shalev-Shwartz, Shai, Gonen, Alon, and Shamir, Ohad. Large-scale convex minimization with a low-rank constraint. In *ICML*, 2011.

Srebro, Nathan, Rennie, Jason D. M., and Jaakkola, Tommi. Maximum-margin matrix factorization. In *NIPS*, 2004.

Sturm, Jos F. Using sedumi 1.02, a MATLAB toolbox for optimization over symmetric cones. *Optimization Methods and Software*, 11–12:625–653, 1999.

Weinberger, Kilian Q., Sha, Fei, Zhu, Qihui, and Saul, Lawrence K. Graph laplacian regularization for large-scale semidefinite programming. In *NIPS*, 2006.

Wen, Zaiwen, Goldfarb, Donald, and Yin, Wotao. Alternating direction augmented lagrangian methods for semidefinite programming. *Math. Prog. Comp.*, 2(3-4): 203–230, 2010.

Xing, Eric P., Ng, Andrew Y., Jordan, Michael I., and Russell, Stuart. Distance metric learning, with application to clustering with side-information. In *NIPS*, 2002.